\def\year{2020}\relax
\documentclass[letterpaper]{article} 
\usepackage{aaai20}  
\usepackage{times}  
\usepackage{helvet} 
\usepackage{courier}  
\usepackage[hyphens]{url}  
\usepackage{graphicx} 
\usepackage{amsmath}
\usepackage{booktabs}
\usepackage{algorithm}
\usepackage{algorithmic}
\usepackage{multirow}
\usepackage{amssymb}
\usepackage{array}
\usepackage{bm}
\usepackage{verbatim}
\usepackage{color}
\usepackage{booktabs}
\usepackage{graphicx}  
\title{Learning to Select Bi-Aspect Information for Document-Scale \\Text Content Manipulation }

\author{Xiaocheng Feng, \textsuperscript{\rm 1}
Yawei Sun, \textsuperscript{\rm 1}
Bing Qin, \textsuperscript{\rm 1}
Heng Gong, \textsuperscript{\rm 1}
Yibo Sun, \textsuperscript{\rm 1} \\
\Large \textbf{
Wei Bi, \textsuperscript{\rm 2}
Xiaojiang Liu, \textsuperscript{\rm 2}
Ting Liu \textsuperscript{\rm 1}}\\
\textsuperscript{\rm 1}Harbin Institute of Technology, Harbin, China \\
\textsuperscript{\rm 2}Tencent AI Lab, Shenzhen, China \\
 \{xcfeng, ywsun, qinb, hgong, ybsun, tliu\}@ir.hit.edu.cn ~~ \{victoriabi,  kieranliu\}@tencent.com
 }

\begin{document}

\maketitle

\begin{abstract}

In this paper, we focus on a new practical task,  document-scale text content manipulation, which is the opposite of text style transfer and aims to preserve text styles while altering the content.
In detail, the input is a set of structured records and a reference text for describing another recordset. 
The output is a summary that accurately describes the partial content in the source recordset with the same writing style of the reference.
The task is unsupervised due to lack of parallel data, and is challenging to select suitable records and style words from bi-aspect inputs respectively and generate a high-fidelity long document.
To tackle those problems, we first build a dataset based on a basketball game report corpus as our testbed, and present an unsupervised neural model with interactive attention mechanism, which is used for learning the semantic relationship between records and reference texts to achieve better content transfer
and better style preservation.
In addition, we also explore the effectiveness of the back-translation in our task for constructing some pseudo-training pairs.
Empirical results show superiority of our approaches over competitive methods, and the models also yield a new state-of-the-art result on a sentence-level dataset.
\footnote{Our code and data are available at: https://github.com/syw1996/SCIR-TG-Data2text-Bi-Aspect}

\end{abstract}

  \begin{figure}[t]
   \centering
   \begin{center}
     \includegraphics*[width=1\linewidth]{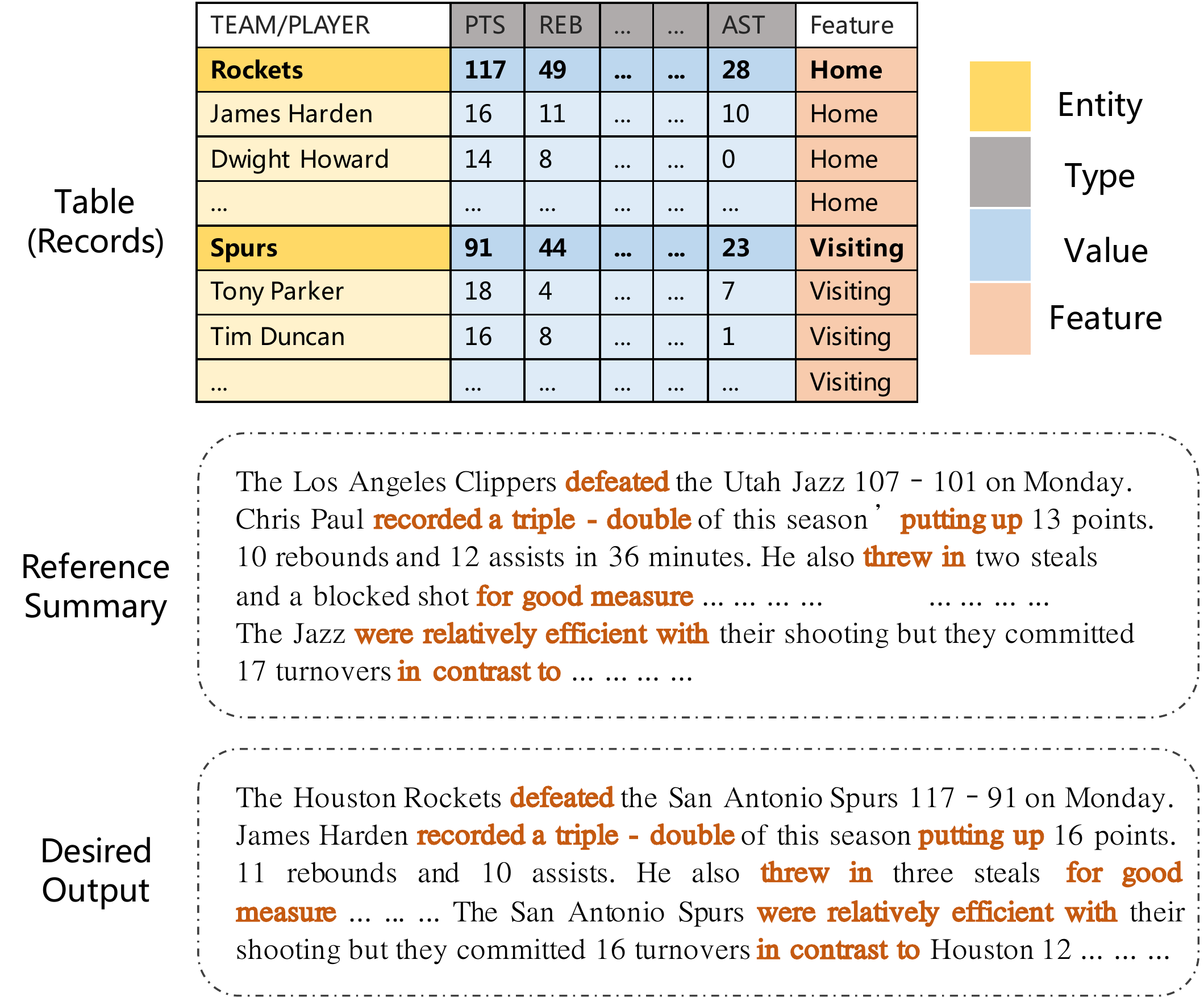}
   \caption{An example input (Table and Reference Summary) of document-level text content manipulation and its desired output. Text portions that fulfill the writing style are highlight in orange.}
   \label{example}
   \vspace{-0.7cm}
   \end{center}
\end{figure}

\section{Introduction}


Data-to-text generation is an effective way to solve data overload, especially with the development of sensor and data storage technologies, which have rapidly increased the amount of data produced in various fields such as weather, finance, medicine and sports \cite{barzilay2005collective}. 
However, related methods are mainly focused on content fidelity, ignoring and lacking control over language-rich style attributes \cite{wang2019toward}.
For example, a sports journalist prefers to use some repetitive words when describing different games \cite{iso2019learning}. 
It can be more attractive and practical to generate an article with a particular style that is describing the conditioning content.

In this paper, we focus on a novel research task in the field of text generation, named document-scale text content manipulation.
It is the task of converting contents of a document into another while preserving the content-independent style words.
For example, given a set of \textbf{structured records} and a \textbf{reference report}, such as statistical tables for a basketball game and a summary for another game, we aim to automatically select partial items from the given records and describe them with the same writing style (e.g., logical expressions, or wording, transitions) of the reference text to directly generate a new report (Figure 1).

\begin{table*}[h]
	\begin{center}
	 \begin{tabular}{lccc}
			\toprule[1pt]
			{}& Train(D/S) &  Dev(D/S) & Test(D/S)   \\
            \hline
            \#Instances & 3371/31,751 & 722/6,833 & 728/6,999  \\
            Avg Ref Length & 335.55/25.90 & 341.17/25.82 & 346.83/25.99  \\
            \#Data Types & 37/34 & 37/34 & 37/34 \\
            Avg Input Record Length & 606/5  & 606/5  & 606/5 \\
            Avg Output Record Length & 38.05/4.88  & 37.80/4.85  & 31.32/4.94 \\
            \bottomrule[1pt]
	\end{tabular}
	\end{center}
	\vspace{-0.2cm}
    \caption{Document-level/Sentence-level Data Statistics.}
    \vspace{-0.4cm}
	\label{statistic}
\end{table*}

In this task, the definition of the text content (e.g., statistical records of a basketball game) is clear, but the text style is vague \cite{dai2019style}. It is difficult to construct paired sentences or documents for the task of text content manipulation.
Therefore, the majority of existing text editing studies develop \textit{controlled} generator with unsupervised generation models, such as Variational Auto-Encoders (VAEs) \cite{kingma2013auto}, Generative Adversarial Networks (GANs) \cite{goodfellow2014generative} and auto-regressive networks \cite{oord2016pixel} with additional pre-trained discriminators.

Despite the effectiveness of these approaches, it remains challenging to generate a high-fidelity long summary from
the inputs. 
One reason for the difficulty is that the input structured records for document-level generation are complex and redundant to determine which part of the data should be mentioned based on the reference text.
Similarly, the model also need to select the suitable style words according to the input records.
One straightforward way to address this problem is to use the relevant algorithms in data-to-text generation, such as pre-selector \cite{mei2015talk} and content selector \cite{puduppully2018data}.
However, these supervised methods cannot be directly transferred considering that we impose an additional goal of preserving the style words, which lacks of parallel data and explicit training objective.
In addition, when the generation length is expanded from a sentence to a document, the sentence-level text content manipulation method \cite{wang2019toward} can hardly preserve the style word (see case study, Figure 4).


In this paper, we present a neural encoder-decoder architecture to deal with document-scale text content manipulation.
In the first, we design a powerful hierarchical record encoder to model the \textbf{structured records}.
Afterwards, instead of modeling records and reference summary as two independent modules \cite{wang2019toward}, we create fusion representations of records and reference words by an interactive attention mechanism. 
It can capture the semantic relatedness of the source records with the reference text to enable the system with the capability of content selection from two different types of inputs.
Finally, we incorporate back-translation \cite{sennrich2016improving} into the training procedure to further improve results, which provides an extra training objective for our model.

To verify the effectiveness of our text manipulation approaches, we first build a large unsupervised document-level text manipulation dataset, which is extracted from an NBA game report corpus \cite{wiseman2017challenges}.
Experiments of different methods on this new corpus show that our full model achieves 35.02 in Style BLEU and 39.47 F-score in Content Selection, substantially better than baseline methods.
Moreover, a comprehensive evaluation with human judgment demonstrates that integrating interactive attention and back-translation could improve the content fidelity and  style preservation of summary by a basic text editing model.
In the end, we conduct extensive experiments on a sentence-level text manipulation dataset \cite{wang2019toward}.
Empirical results also show that the proposed approach achieves a new state-of-the-art result.


\section{Preliminaries}

\subsection{Problem Statement}

Our goal is to automatically select partial items from the given content and describe them with the same writing style of the reference text.
 As illustrated in Figure 1, each input instance consists of a statistical table $x$ and a reference summary $y'$.
 We regard each cell in the table as a record  $r=\{r_{o}\}_{o=1}^{L_x}$, where $L_x$ is the number of records in table $x$.
 Each record $r$ consists of four types of information including entity $r.e$ (the name of team or player, such as LA Lakers or Lebron James), type $r.t$ (the types of team or player, e.g., points, assists or rebounds) and value $r.v$ (the value of a certain player or team on a certain type), as well as feature $r.f$ (e.g., home or visiting) which indicates whether a player or a team compete in home court or not. 
 In practice, each player or team takes one row in the table and each column contains a type of record such as points, assists, etc.
 The reference summary or report consists of multiple sentences, which are assumed to describe content that has the same types but different entities and values with that of the table $x$.
 
 Furthermore, following the same setting in sentence-level text content manipulation \cite{wang2019toward}, we also provide additional information at training time. 
 For instance, each given table $x$ is paired with a corresponding $y_{aux}$, which was originally written to describe $x$ and each reference summary $y'$ also has its corresponding table $x'$ containing the records information.
 The additional information can help models to learn the table structure and how the desired records can be expressed in natural language when training.
 It is worth noting that we do not utilize the side information beyond $(x, y')$ during the testing phase and the task is unsupervised as there is no ground-truth target text for training.

\subsection{Document-scale Data Collection}

 In this subsection, we construct a large document-scale text content manipulation dataset as a testbed of our task. 
 The dataset is derived from an NBA game report corpus ROTOWIRE \cite{wiseman2017challenges}, which consists of 4,821 human written NBA basketball game summaries aligned with their corresponding game tables.
 In our work, each of the original table-summary pair is treated as a pair of $(x, y_{aux})$, as described in previous subsection.
 To this end, we design a \textbf{type-based} method for obtaining a suitable reference summary $y'$ via retrieving another table-summary from the training data using $x$ and $y_{aux}$.
 The retrieved $y'$ contains record types as same as possible with record types contained in $y$.
 We use an existing information extraction tool \cite{wiseman2017challenges} to extract record types from the reference text. 
 Table \ref{statistic} shows the statistics of constructed document-level dataset and a sentence-level benchmark dataset \cite{wang2019toward}.
 We can see that the proposed document-level text manipulation problem is more difficult than sentence-level, both in terms of the complexity of input records and the length of generated text.

\begin{figure}[h]

   \centering
   \begin{center}
   \vspace{-0.2cm}
     \includegraphics*[width=1\linewidth]{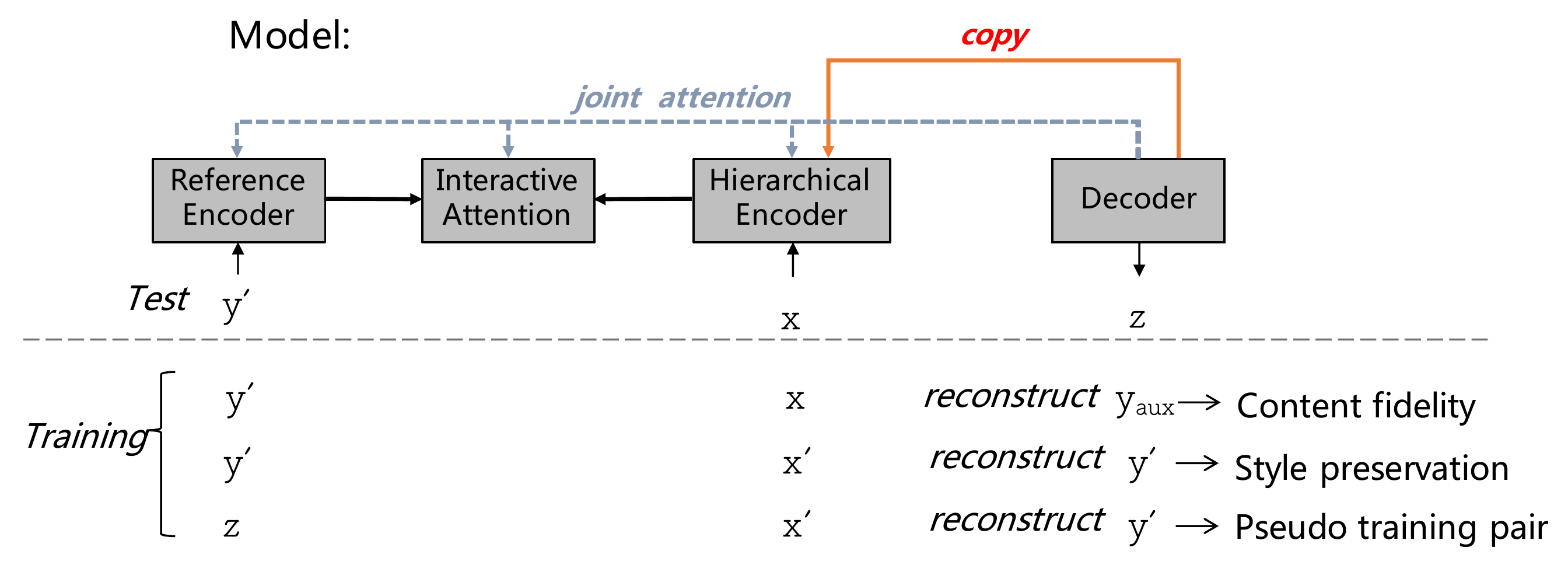}
     \vspace{-0.4cm}
   \caption{An overview of the document-level approach.}
   \label{example}
   \vspace{-0.6cm}
   \end{center}
\end{figure}

\section{The Approach}

\begin{figure*}[!htb]
	\centering
	\includegraphics[width=0.95\textwidth]{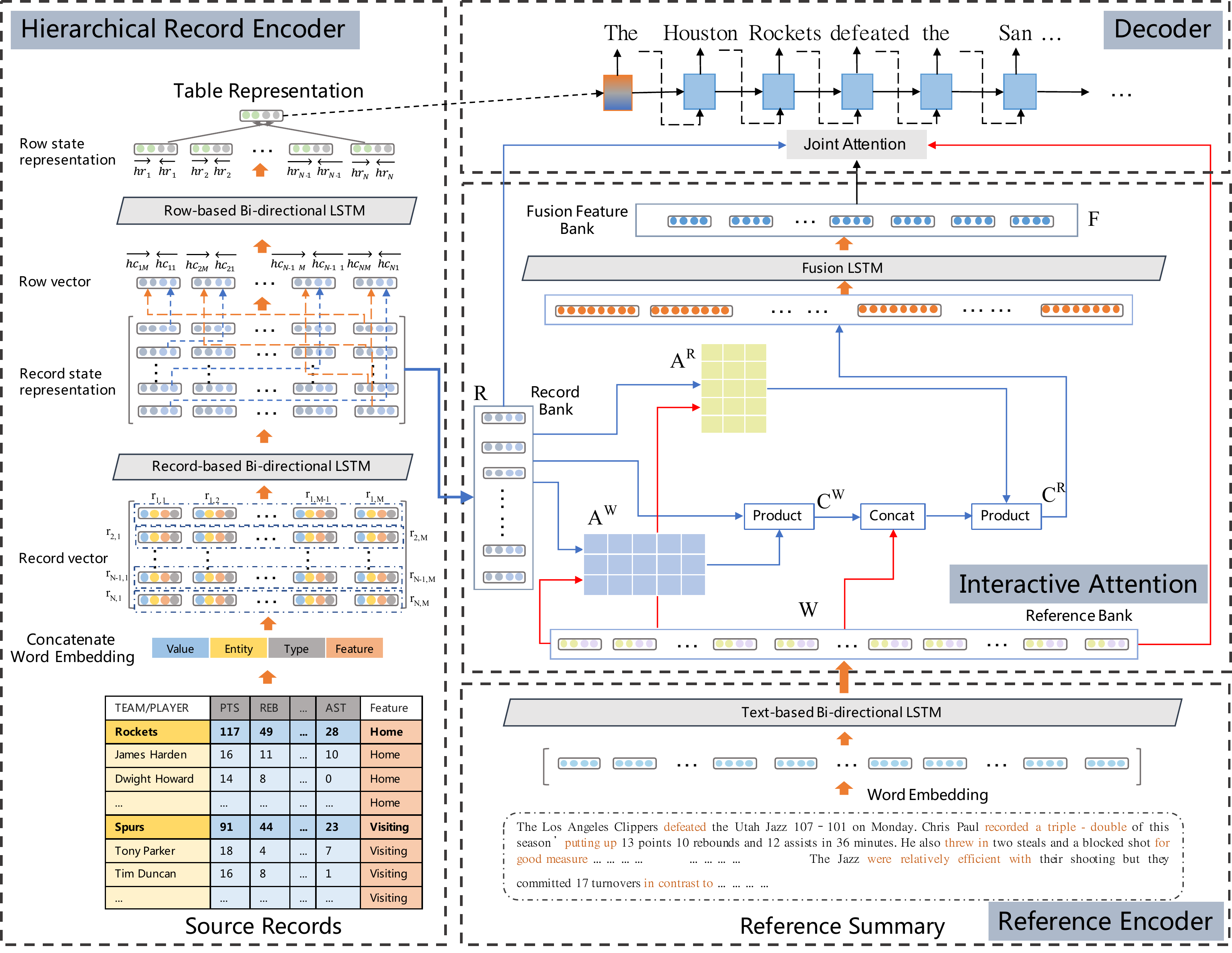}
    \vspace{-0.2cm}
            \caption{The architecture of our proposed model.}
          \label{modelstructure} 
          \vspace{-0.4cm}

\end{figure*}

This section describes the proposed approaches to tackle the document-level problem.
We first give an overview of our architecture.
Then, we provide detailed formalizations of our model with special emphasize on Hierarchical Record Encoder, Interactive Attention, Decoder and Back-translation.

\subsection{An Overview}

In this section, we present an overview of our model for document-scale text content manipulation, as illustrated in Figure 2.
Since there are unaligned training pairs, the model is trained with three competing objectives of reconstructing the auxiliary document $y_{aux}$ based on $x$ and $y'$ (for content fidelity), the reference document $y'$ based on $x'$ and $y'$ (for style preservation), and the reference document $y'$ based on $x'$ and pseudo $z$ (for pseudo training pair).
Formally, let $p_{\theta}=(z|x,y')$ denotes the model that takes in records $x$ and a reference summary $y'$, and generates a summary $z$. 
Here $\theta$ is the model parameters.
In detail, the model consists of a reference encoder, a record encoder, an interactive attention and a decoder.

The first reference encoder is used to extract the representation of reference summary $y'$ by employing a bidirectional-LSTM model \cite{hochreiter1997long}.
The second record encoder is applied to learn the representation of all records via hierarchical modeling on record-level and row-level.
The interactive attention is a co-attention method for learning the semantic relationship between the representation of each record and the representation of each reference word.
The decoder is another LSTM model to generate the output summary with a hybrid attention-copy mechanism at each decoding step.

Note that we set three goals, namely content fidelity, style preservation and pseudo training pair.
Similar to sentence-scale text content manipulation \cite{wang2019toward}, the first two goals are simultaneous and in a sense competitive with each other (e.g., describing the new designated content would usually change the expressions in reference sentence to some extent).
 The content fidelity objective $L_{record}(\theta)$  and style preservation objective $L_{style}(\theta)$ are descirbed in following equations.
  \begin{align}
L_{record}(\theta)=log_{p_{\theta}(y_{aux}|x,y')}\\
L_{style}(\theta)=log_{p_{\theta}(y'|x',y')}
\end{align}
The third objective is used for training our system in a true text manipulation setting.
We can regard this as an application of the \textbf{back-translation} algorithm in document-scale text content manipulation.
Subsection "Back-translation Objective" will give more details.




\subsection{Hierarchical Record Encoder}

We develop a hierarchical table encoder to model game statistical tables on record-level and row-level in this paper.
It can model the relatedness of a record with other records in same row and a row (e.g., a player) with other rows (e.g., other players) in same table.
As shown in the empirical study (see Table 2), the hierarchical encoder can gain significant improvements compared with the standard MLP-based data-to-text model \cite{wiseman2017challenges}.
Each word and figure are represented as a low dimensional, continuous and real-valued vector, also known as word embedding \cite{mikolov2013distributed,pennington2014glove}.
All vectors are stacked in a word embedding matrix $L_w \in \mathbb{R}^{d \times |V|}$, where $d$ is the dimension of the word vector and $|V|$ is the vocabulary size.

On record-level, we first concatenate the embedding of record's entity, type, value and feature as an initial representation of the record  ${{r_{ij}}} = \{{r_{ij}.e};{r_{ij}.t};{r_{ij}.v};{r_{ij}.f} \}\in \mathbb{R}^{4d \times 1} $, where ${i, j}$ denotes a record in the table of $i^{th}$ row and $j^{th}$ column as mentioned in Section 2.1.
Afterwards, we employ a bidirectional-LSTM to model records of the same row.
For the $i^{th}$ row, we take record  $\{ r_{i1}, ...,r_{ij}, ..., r_{iM} \}$ as input, then obtain record's forward hidden representations $\{ \overrightarrow{hc_{i1}}, ...,\overrightarrow{hc_{ij}}, ..., \overrightarrow{hc_{iM}} \}$ and backward hidden representations $\{ \overleftarrow{hc_{i1}}, ...,\overleftarrow{hc_{ij}}, ..., \overleftarrow{hc_{iM}} \}$, where $M$ is the number of columns (the number of types). 
In the end, we concatenate $\overrightarrow{hc_{ij}}$ and $\overleftarrow{hc_{ij}} $  as a final representation of record $r_{ij}$ and concatenate $\overrightarrow{hc_{iM}}$ and $\overleftarrow{hc_{i1}}$  as a hidden vector of the $i^{th}$ row.

On row-level, the modeled row vectors are fed to another bidirectional-LSTM model to learn the table representation.
In the same way, we can obtain row's forward hidden representations $\{ \overrightarrow{hr_{1}}, ...,\overrightarrow{hr_{i}}, ..., \overrightarrow{hr_{N}} \}$ and backward hidden representations $\{ \overleftarrow{hr_{1}}, ...,\overleftarrow{hr_{i}}, ..., \overleftarrow{hr_{N}} \}$, where $N$ is the number of rows (the number of entities). 
And the concatenation of $[\overrightarrow{hr_{i}}, \overleftarrow{hr_{i}}]$ is regarded as a final representation of the $i^{th}$ row.
An illustration of this network is given in the left dashed box of Figure 3, where the two last hidden vector $\overrightarrow{hr_{N}}$ and $\overleftarrow{hr_{1}}$ can be concatenated as the table representation, which is the initial input for the decoder.

Meanwhile, a bidirectional-LSTM model is used to encode the reference text $ {w_1, ..., w_K}$ into a set of hidden states $W = [{w.h_1, ..., w.h_K}]$, where $K$ is the length of the reference text and each $w.h_i$ is a $2d$-dimensional vector.

\subsection{Interactive Attention}

We present an interactive attention model that attends to the structured records and reference text simultaneously, and finally fuses both attention context representations.
Our work is partially inspired by the successful application of co-attention methods in Reading Comprehension \cite{cui2016attention,xiong2016dynamic,yin2018zero} and Natural Language Inference \cite{duan2018attention,conneau2017supervised}.

As shown in the middle-right dashed box of Figure 3,
we first construct the Record Bank as $R= [rc_1,...,rc_o,..., rc_{L_x},] \in \mathbb{R}^{2d \times L_x}$,  where $L_x = M \times N$ is the number of records in Table $x$ and each $rc_o$ is the final representation of record $r_{ij}$, $r_{ij} = [\overrightarrow{hc_{ij}}, \overleftarrow{hc_{ij}}]$, as well as the Reference Bank $W$, which is $W  = [{w.h_1, ..., w.h_K}] $.
Then, we calculate the affinity matrix, which contains affinity scores corresponding to all pairs of structured records and reference words: $L = R^TW \in \mathbb{R}^{ L_x \times K} $.
The affinity matrix is normalized row-wise to produce the attention weights $A^W$ across the structured table for each word in the reference text, and column-wise to produce the attention weights $A^R$ across the reference for each record in the Table:
  \begin{align}
A^W=softmax(L)\in \mathbb{R}^{ L_x \times K}  \\
A^R=softmax(L^T)\in \mathbb{R}^{K \times L_x} 
\end{align}
Next, we compute the suitable records of the table in light of each word of the reference.
\begin{equation}
C^W=  R A^W \in \mathbb{R}^{ 2d \times K}  
\end{equation}
We similarly compute the summaries $WA^R$ of the reference in light of each record of the table.
Similar to \cite{cui2016attention}, we also place reference-level attention over the record-level attention by compute the record summaries $C^WA^R$ of the previous attention weights in light of each record of the table.
These two operations can be done in parallel, as is shown in Eq. 6.
\begin{equation}
C^R= [W; C^W]A^R  \in \mathbb{R}^{ 4d \times  L_x}  
\end{equation}
We define $C^R$ as a fusion feature bank, which is an interactive representation of the reference and structured records.

In the last, a bidirectional LSTM is used for fusing the relatedness to the interactive features.
The output $F = [f_1,..., f_{L_X}] \in \mathbb{R}^{ 2d \times  L_x} $, which provides a foundation for selecting which record may be the best suitable content, as fusion feature bank. 

\subsection{Decoder}

An illustration of our decoder is shown in the top-right dashed box of Figure 3.
We adopt a joint attention model \cite{luong2015effective} and a copy mechanism \cite{gu2016incorporating} in our decoding phrase.
In particular, our joint attention covers the fusion feature bank,  which represents an interactive representation of the input records and reference text.
And we refuse the coverage mechanism, which does not satisfy the original intention of content selection in our setting.

In detail, we present a flexible copying mechanism which is able to copy contents from table records.
The basic idea of the copying mechanism is to copy a word from the table contents as a trade-off of generating a word from target vocabulary via softmax operation.
On one hand, we define the probability of copying a word $\tilde{z}$ from table records at time step $t$ as $g_t(\tilde{z}) \odot \alpha_{(t, id(\tilde{z}))}$, where $g_t(\tilde{z})$ is the probability of copying a record from the table, $id(\tilde{z})$ indicates the record number of $\tilde{z}$, and $\alpha_{(t, id(\tilde{z}))}$ is the attention probability on the $id(\tilde{z})$-th record.
On the other hand, we use $(1 - g_t(\tilde{z}) ) \odot \beta_{(\tilde{z})}$ as the
probability of generating a word $\tilde{z}$ from the target vocabulary, where $\beta_{(\tilde{z})}$ is from the distribution over the target vocabulary via softmax operation.
We obtain the final probability of generating a word $\tilde{z}$ as follows
\begin{equation}
P_t(\tilde{z}) = g_t(\tilde{z}) \odot \alpha_{(t, id(\tilde{z}))} + (1 - g_t(\tilde{z}) ) \odot \beta_{(\tilde{z})}
\end{equation}
The above model, copies contents only from table records, but not reference words.

\subsection{Back-translation Objective}

In order to train our system with a true text manipulation setting, we adapt the back-translation \cite{sennrich2016improving} to our scenario.
After we generate text $z$ based on $(x, y')$, we regard $z$ as a new reference text and paired with $x'$ to generate a new text $z'$.
Naturally, the golden text of  $z'$ is $y'$, which can provide an additional training objective in the training process.
Figure 2 provides an illustration of the back-translation, which reconstructs $y'$ given ($x'$, $z$):
\begin{equation}
L_{backtrans}(\theta)=log_{p_{\theta}(y'|x',z)}
\end{equation}
We call it the back-translation objective.
Therefore, our final objective consists of content fidelity objective, style preservation objective and back-translation objective.
\begin{equation}
\begin{split}
L_{joint}(\theta)=&\lambda_1 L_{record}(\theta) + \lambda_2 L_{style}(\theta) \\
& + (1 - \lambda_1 - \lambda_2) L_{back-trans}(\theta)
\end{split}
\label{final_loss}
\end{equation}
where $\lambda_1 $ and $\lambda_2$ are hyperparameters.

\section{Experiments}

In this section, we describe experiment settings and report the experiment results and analysis.
We apply our neural models for text manipulation on both document-level and sentence-level datasets, which are detailed in Table 1.

\subsection{Implementation Details and Evaluation Metrics}

We use two-layers LSTMs in all encoders and decoders, and employ  attention mechanism  \cite{luong2015effective}. Trainable model parameters are randomly initialized under a Gaussian distribution. We set the hyperparameters empirically based on multiple tries with different settings. 
We find the following setting to be the best. The dimension of word/feature embedding, encoder hidden state, and decoder hidden state are all set to be 600. We apply dropout at a rate of 0.3.
Our training process consists of three parts. 
In the first, we set $\lambda_1=0$ and $\lambda_2=1$ in Eq. 7 and pre-train the model to convergence. 
We then set $\lambda_1=0.5$ and $\lambda_2=0.5$ for the next stage training. 
Finally, we set $\lambda_1=0.4$ and $\lambda_2=0.5$ for full training. Adam is used for parameter optimization with an initial learning rate of 0.001 and decaying rate of 0.97. 
During testing, we use beam search with beam size of 5. The minimum decoding length is set to be 150 and maximum decoding length is set to be 850.

We use the same evaluation metrics employed in \cite{wang2019toward}.
\textbf{Content Fidelity (CF)} is an information extraction (IE) approach used in \cite{wiseman2017challenges} to measure model's ability to generate text containing factual records.
That is, precision and recall (or number) of unique records extracted from the generated text $z$ via an IE model also appear in source recordset $x$.
\textbf{Style Preservation} is used to measure how many stylistic properties of the reference are retained in the generated text. 
In this paper, we calculate BLEU score between the generated text and the reference to reflect model's ability on style preservation.
Furthermore, in order to measure model's ability on content selection, we adopt another IE-based evaluation metric, named \textbf{Content selection, (CS)}, which is used for data-to-text generation \cite{wiseman2017challenges}.
It is measured in terms of precision and recall by comparing records in generated text $z$ with records in the auxiliary reference $y_{aux}$.

We compare with the following baseline methods on the document-level text manipulation.

(1) Rule-based Slot Filling Method (\textit{Rule-SF}) is a straightforward way for text manipulation. Firstly, It masks the record information $x'$ in the $y'$ and build a mapping between $x$ and  $x'$ through their data types.
Afterwards, select the suitable records from $x$ to fill in the reference y with masked slots.
The method is also used in sentence-level task \cite{wang2019toward}.

(2) Copy-based Slot Filling Method (\textit{Copy-SF}) is a data-driven slot filling method.
It is derived from \cite{li2018point}, which first generates a template text with data slots to be filled and then leverages a delayed copy mechanism to fill in the slots with proper data records.

(3) Conditional Copy based Data-To-Text (\textit{CCDT}) is a classical neural model for data-to-text generation \cite{wiseman2017challenges}.

(4) Hierarchical Encoder for Data-To-Text (\textit{HEDT}) is also a data-to-text method, which adopts the same hierarchical encoder in our model.

(5) Text Manipulation with Table Encoder (\textit{TMTE}) extends sentence-level text editing method \cite{wang2019toward} by equipping a more powerful hierarchical table encoder.

(6) Co-attention-based Method (\textit{Coatt}): a variation of our model by replacing interactive attention with another co-attention model \cite{ji2018incorporating}.

(7) Ours w/o Interactive Attention (\textit{-InterAtt}) is our model without interactive attention.

(8) Ours w/o Back-translation (\textit{-BackT})  is also a variation of our model by omitting back-translation loss.

In addition, for sentence-level task, we adopt the same baseline methods as the paper \cite{wang2019toward}, including an attention-based Seq2Seq method with copy mechanism \cite{sutskever2014sequence}, a rule-based method, two style transfer methods, \textit{MAST} \cite{lample2018multiple} and \textit{AdvST} \cite{logeswaran2018content}, as well as their state-of-the-art method, abbreviate as S-SOTA.  

\subsection{Comparison on Document-level Text Manipulation}

\begin{table}[t]
	\begin{center}
	\fontsize{8pt}{10pt}\selectfont
		\begin{tabular}{p{1.45cm} p{0.5cm}<{\centering}p{0.5cm}<{\centering}p{0.5cm}<{\centering}  p{0.5cm}<{\centering}p{0.5cm}<{\centering} p{0.5cm}<{\centering}  }
			\toprule[1pt]
			\centering
			\multirow{1}{*}[ - 0.2cm]{Model} & \multicolumn{2}{c}{CF} & \multicolumn{3}{c}{CS} & \multicolumn{1}{c}{Style}\\
			
			& P\% &\# & P\% & R\% & F\% & BLEU \\
			
			\hline
			\textit{Rule-SF}				& 53.42  & 7.97  & 26.89  & 22.41 & 24.45  &  \bf{100.00}  \\
			\textit{Copy-SF}		    &75.65& 9.77   & 42.44   & 36.06   &  38.99 & \bf{100.00}   \\
			\hline
			\hline
			\textit{CCDT}		&  75.62 & 22.32  & 32.80  & 39.93  & 36.02  & 15.81 \\
			\textit{HEDT}		& \bf{91.59}   & \bf{32.56}  & 31.62 & \bf{44.22} &36.87 & 17.43 \\
		    \hline
		    \hline
			\textit{TMTE} &  71.97 & 9.58  & 44.48   & 31.50  & 36.88  & 15.55 \\
			\textit{Coatt}		    & 63.08   & 8.30  & 41.89  & 30.16  & 35.07  & 25.93 \\
			\hline
			\hline
		    \textit{Our model}		    & 70.24   & 10.14  & 45.55  & 34.82  &\bf{39.47} & \bf{35.02}\\
		    \textit{ \ w/o InterAtt}		    & {75.76}   & 8.74  & \bf{49.17}  & 30.93  & 37.97& 15.87 \\
		    \textit{ \ w/o BackT}		    & 63.50   & 8.99  & 41.64  & 31.95  & 36.16  & 32.72 \\

			\bottomrule[1pt]
		\end{tabular}
	\end{center}
    \vspace{-0.4cm}
	\caption{\label{Results}  Document-level comparison results.
	\vspace{-0.2cm}
	}
    \vspace{-0.4cm}
\end{table}

Document-level text manipulation experimental results are given in Table 2.
The first block shows two slot filling methods, which can reach the maximum BLEU (100) after masking out record tokens.
It is because that both methods only replace records without modifying other parts of the reference text.
Moreover, Copy-SF achieves reasonably good performance on multiple metrics, setting a strong baseline for content fidelity and content selection.
For two data-to-text generation methods CCDT and HEDT, the latter one is consistently better than the former, which verifies the proposed hierarchical record encoder is more powerful.
However, their Style BLEU scores are particularly low, which demonstrates that direct supervised learning is incapable of controlling the text expression.
In comparison, our proposed models achieve better Style BLEU and Content Selection F\%.
The superior performance of our full model compared to the variant ours-w/o-InterAtt, TMTE and Coatt demonstrates the usefulness of the interactive attention mechanism.

\begin{table}[h]
	\begin{center}
	\fontsize{8pt}{10pt}\selectfont
		\begin{tabular}{p{1.45cm} p{1cm}<{\centering}p{1cm}<{\centering} p{1cm}<{\centering} p{1cm}<{\centering} }
			\toprule[1pt]
			\centering
			\multirow{1}{*}[ - 0.15cm]{Model} & \multicolumn{1}{c}{Content} & \multicolumn{1}{c}{ Style} & \multirow{1}{*}[ - 0.15cm]{Fluency}& \multirow{1}{*}[ - 0.15cm]{Average}\\
			
			& Fidelity  &Preservation &  \\
			
			\hline
			\textit{Rule-SF}				& 1.81  & 4.50  & 2.78  & 3.03  \\
			\textit{Copy-SF}		    & 3.54   & 4.50   & 2.49  & 3.51   \\
			
			\textit{HEDT}		& 3.67   & 1.96  & 3.19  & 2.94 \\
		    
			\textit{TMTE} &   2.36 & 1.12  & 1.95  & 1.81 \\
			
			\hline
		    \textit{Our model}		    & 3.62   & 2.92  & 4.59 & 3.71 \\

			\bottomrule[1pt]
		\end{tabular}
	\end{center}
    \vspace{-0.4cm}
	\caption{\label{Results} Human Evaluation Results. 
	}
    \vspace{-0.4cm}
\end{table}

\subsection{Human Evaluation}

In this section, we hired three graduates who passed intermediate English test (College English Test Band 6) and were familiar with NBA games to perform human evaluation.
Following \cite{wang2019toward,shen2017style}, we presented to annotators five generated summaries, one from our model and four others from comparison methods, such as \textit{Rule-SF}, \textit{Copy-SF}, \textit{HEDT}, \textit{TMTE}.
These students were asked to rank the five summaries by considering  “Content Fidelity”, “Style  Preservation” and “Fluency” separately.
The rank of each aspect ranged from 1 to 5 with the higher score the better and the ranking scores are averaged as the final score.
For each study, we evaluated on 50 test instances.
From Table 3, we can see that the Content Fidelity and Style Preservation results are highly consistent with the results of the objective evaluation.
An exception is that the Fluency of our model is much higher than other methods.
One possible reason is that the reference-based generation method is more flexible than template-based methods, and more stable than pure language models on document-level long text generation tasks.

\subsection{Comparison on Sentence-level Text Manipulation}

\begin{table}[t]
	\begin{center}
	\fontsize{8pt}{10pt}\selectfont
		\begin{tabular}{p{1.60cm} p{1cm}<{\centering}p{1cm}<{\centering} p{1cm}<{\centering} p{1cm}<{\centering}  }
			\toprule[1pt]
			\centering
			\multirow{1}{*}[ - 0.2cm]{Model} & \multicolumn{3}{c}{Content Fideity} &  \multicolumn{1}{c}{Style}\\
			\cline{2-4} 
			
			& Percision\% &Recall\% &F1\% & BLEU \\
			
			\hline
			\textit{Rule-based}				& 62.63  & 63.64 & 63.13 & \bf{100.00}    \\
			\textit{AttnCopy-S2S}		    & \bf{88.71}   & 60.64 & 72.04  & 39.15    \\
			\hline
			\hline
			\textit{MAST}		&  33.15 & 31.09 & 32.09 & \bf{95.29}   \\
			\textit{AdvST}		& 66.51   & 56.03 & 60.82  & 72.22  \\
			\textit{S-SOTA}		    & 78.31   & 65.64 & 71.42 & 80.83 \\
		    \hline
			\hline
		    \textit{Our model}		    &   74.41 & \bf{79.43} & \bf{76.84} & 81.92  \\
		    \textit{ \  w/o InterAtt}		    & 74.48   & 79.12 & 76.73 & 81.06   \\
		    \textit{ \  w/o BackT}		    & 73.42   & 77.83 & 75.56 & 79.88   \\

			\bottomrule[1pt]
		\end{tabular}
	\end{center}
    \vspace{-0.4cm}
	\caption{\label{Results} Sentence-level comparison results.
	}
    \vspace{-0.4cm}
\end{table}

To demonstrate the effectiveness of our models on sentence-level text manipulation, we show the results in Table 4. 
We can see that our full model can still get consistent improvements on  sentence-level task over previous state-of-the-art method. 
Specifically, we observe that interactive attention and back-translation cannot bring a significant gain.
This is partially because the input reference and records are relatively simple, which means that they do not require overly complex models for representation learning.

\subsection{Qualitative Example}

\begin{figure*}[!htb]
	\centering
	\includegraphics[width=0.98\textwidth]{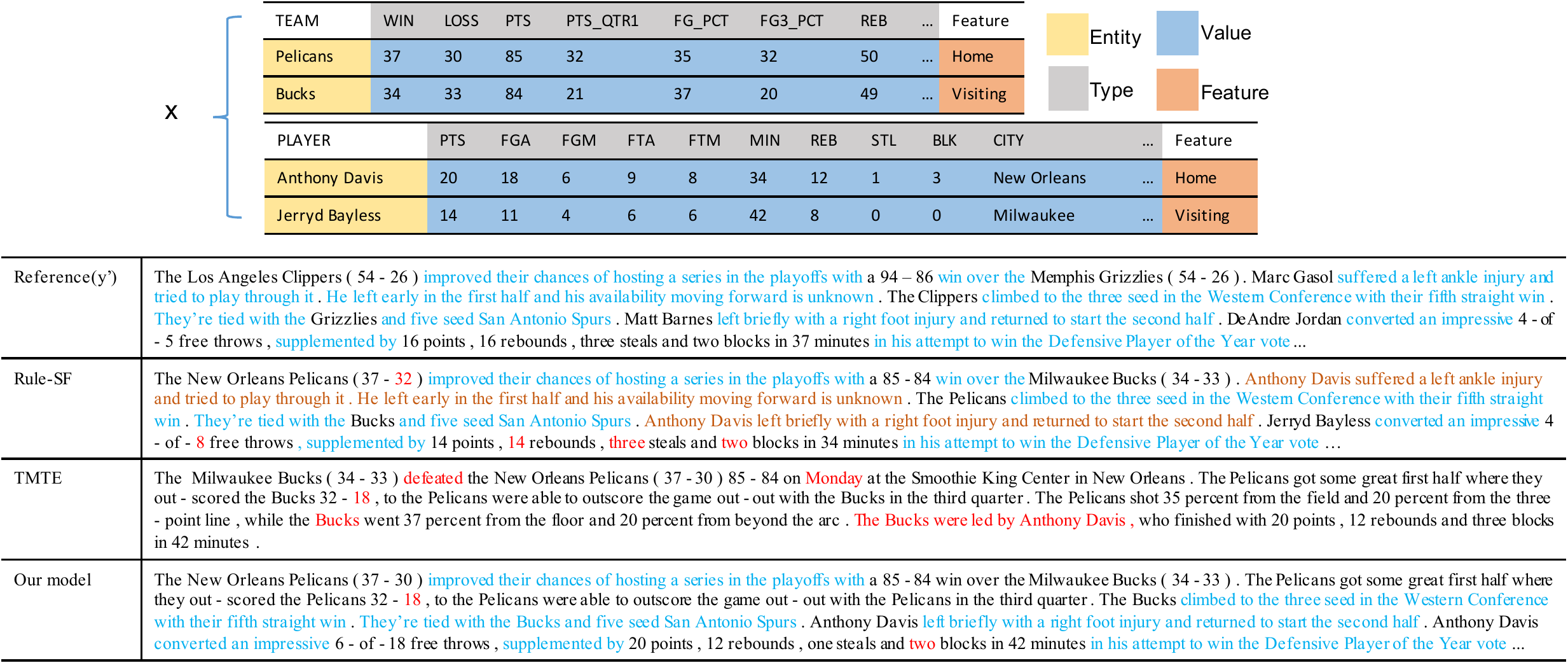}
            \caption{Examples of model output for HEDT, Rule-SF and Our full model on document-scale dataset. Red words or numbers are fidelity errors. Orange words are logical errors. Text portions in the reference summary and the document-scale outputs of different generation model that fulfill the stylistic characteristics are highlighted in blue.}
          \label{modelstructure} 
          \vspace{-0.4cm}

\end{figure*}

Figure 4 shows the generated examples by different models given content records $x$ and reference summary $y'$.
We can see that our full model can manipulate the reference style words more accurately to express the new records.
Whereas four generations seem to be fluent, the summary of Rule-SF includes logical erroneous sentences colored in orange.
It shows a common sense error that Davis was injured again when he had left the stadium with an injury.
This is because although the rule-based method has the most style words, they cannot be modified, which makes these style expressions illogical.
An important discovery is that the sentence-level text content manipulation model TMTE fails to generate the style words similar to the reference summary.
The reason is that TMTE has no interactive attention module unlike our model, which models the semantic relationship between records and reference words and therefore accurately select the suitable information from bi-aspect inputs.
However, when expressions such as parallel structures are used, our model generates erroneous expressions as illustrated by the description about Anthony Davis's records ``20 points, 12 rebounds, one steals and two blocks in 42 minutes''.



\section{Related Work}

Recently, text style transfer and controlled text generation have been widely studied \cite{hu2017toward,shen2017style,logeswaran2018content,tian2018structured}.
They mainly focus on generating realistic sentences, whose attributes can be controlled by learning disentangled latent representations.
Our work differs from those in that:
(1) we present a document-level text manipulation task rather than sentence-level.
(2) The style attributes in our task is the textual expression of a given reference document.
(3) Besides text representation learning, we also need to model structured records in our task and do content selection.
Particularly, our task can be regard as an extension of sentence-level text content manipulation \cite{wang2019toward}, which assumes an existing sentence to provide the source of style and structured records as another input.
It takes into account the semantic relationship between records and reference words and experiment results verify the effectiveness of this improvement on both document- and sentence-level datasets.

Furthermore, our work is similar but different from data-to-text generation studies  \cite{mei2015talk,nie2018operations,sha2018order,liu2018table,puduppully2018data,bao2018table,sun2018semantic,gong2019table1,chen2019enhancing,gong2019table2}.
This series of work focuses on generating more accurate descriptions of given data, rather than studying the writing content of control output. 
Our task takes a step forward to simultaneously selecting desired content and depending on specific reference text style.
Moreover, our task is more challenging due to its unsupervised setting.
Nevertheless, their structured table modeling methods and data selection mechanisms can be used in our task.
For example, \cite{wiseman2017challenges} develops a MLP-based table encoder.
\cite{li2018point} presents a two-stage approach with a delayed copy mechanism, which is also used as a part of our automatic slot filling baseline model.

\section{Conclusion}
 
 In this paper, we first introduce a new yet practical problem, named document-level text content manipulation,   which aims to express given structured recordset with a paragraph text and mimic the writing style of a reference text.
Afterwards, we construct a corresponding dataset and develop a neural model for this task with hierarchical record encoder and interactive attention mechanism.
In addition, we optimize the previous training strategy with back-translation.
Finally, empirical results verify that the presented approaches perform substantively better than several popular data-to-text generation and style transfer methods on both constructed document-level dataset and a sentence-level dataset.
In the future, we plan to integrate neural-based retrieval methods into our model for further improving results.

\section*{Acknowledgments}

Bing Qin is the corresponding author of this work.
This work was supported by 
the National Key R\&D Program of
China (No. 2018YFB1005103), National Natural Science
Foundation of China (No. 61906053) and Natural Science Foundation of Heilongjiang Province of China (No. YQ2019F008).


\bibliography{aaai2020}
\bibliographystyle{aaai.bst}

\end{document}